\documentclass{article}

\PassOptionsToPackage{numbers, compress}{natbib}

\usepackage[preprint]{neurips_2026}

\usepackage[utf8]{inputenc} 
\usepackage[T1]{fontenc}    
\usepackage{hyperref}       
\usepackage{url}            
\usepackage{booktabs}       
\usepackage{amsfonts}       
\usepackage{nicefrac}       
\usepackage{microtype}      
\usepackage{xcolor}         
\usepackage{graphicx}       
\usepackage{amsmath}
\usepackage{amsthm}
\usepackage{float}
\usepackage{algorithm}
\usepackage{algorithmic}
\usepackage{multirow}

\newtheorem{theorem}{Theorem}
\newtheorem{lemma}{Lemma}
\newtheorem{proposition}{Proposition}
\newtheorem{assumption}{Assumption}

\title{GIFT: Group-Relative Implicit Fine-Tuning Integrates GRPO with DPO and UNA}

\author{%
  Zhichao Wang \\
  Inflection AI \\
  \texttt{zcwang0201@gmail.com} \\
}

\begin{document}

\maketitle

\begin{abstract}
This paper investigates whether reward matching is a viable alternative to reward maximization methods for on-policy RL of LLMs. Group-relative Implicit Fine-Tuning (GIFT) is proposed, combining GRPO-style group sampling, DPO-style implicit reward, and UNA-style MSE between implicit and explicit advantages. 
By applying z-score standardization, the intractable partition function $Z(x)$ in the DPO implicit reward is canceled, and the KL coefficient $\beta$ is eliminated from the RLHF and RLVR objective.
The population minimizers of $\mathcal{L}_{\text{GIFT}}$ are characterized in closed form: they coincide exactly with the GRPO/RLHF solution family $\pi^{*}_{\beta}(y|x)\propto\pi_{\text{ref}}(y|x)e^{\frac{1}{\beta}r_{\phi}(x,y)}$, with a prompt-dependent, variance-determined KL coefficient $\beta(x)=\frac{\sigma_\phi(x)}{\hat{\sigma}_\theta(x)}$. GIFT therefore solves the same parametric policy family as GRPO while replacing GRPO's externally tuned scalar $\beta$ with a prompt-adaptive $\beta(x)$ optimized endogenously by matching reward distributions.
Empirically, on 7B-32B backbones, GIFT converges faster than GRPO, DAPO and GSPO and overfits less on RLVR (GSM8K, MATH, AIME) and produces higher length-controlled win rates on RLHF (AlpacaEval, Arena-Hard). All proofs and detailed background are deferred to the appendix.
\end{abstract}

\section{Introduction}

Modern LLMs are pretrained on trillions of tokens to acquire broad linguistic and factual knowledge \cite{openai2024gpt4ocard}, after which supervised fine-tuning (SFT) instills instruction-following and dialogue ability \cite{ouyang2022traininglanguagemodelsfollow}. Even after SFT, models still need to be aligned with human preferences and downstream task objectives.
Reinforcement learning-based post-training methods address these issues from several angles. RLHF \cite{ouyang2022traininglanguagemodelsfollow} trains a Bradley-Terry reward model \cite{bradley1952rank} from pairwise preferences and then optimizes the policy by PPO \cite{schulman2017proximalpolicyoptimizationalgorithms} subject to a KL penalty. RLVR \cite{deepseekai2025deepseekr1incentivizingreasoningcapability} replaces learned reward models with rule-based correctness signals; GRPO \cite{shao2024deepseekmathpushinglimitsmathematical} reduces PPO's compute by group-relative reward normalization. In parallel, DPO \cite{rafailov2024directpreferenceoptimizationlanguage} reformulates RLHF as a single-stage offline objective using an implicit reward, and UNA \cite{wang2025unaunifyingalignmentsrlhfppo} extends DPO to pointwise data via an MSE between implicit and explicit rewards. While DPO and UNA are computationally efficient, they lack on-policy exploration and can underperform on tasks where new responses must be discovered.

\paragraph{Question.} Is the clipped policy-gradient surrogate of PPO/GRPO the only useful policy gradient for on-policy LLM post-training, or can a regression-style loss between implicit and explicit rewards--of the kind UNA uses offline--induce a different but equally valid policy gradient on-policy?

\paragraph{Approach.} \textit{Group-relative Implicit Fine-Tuning (GIFT)} samples $N$ on-policy responses per prompt, computes a DPO-style implicit reward and a (verifiable or learned) explicit reward for each, and applies the same group $z$-score standardization to both. The $z$-score of a reward is, by definition, an \emph{advantage}: the explicit reward $r_\phi$ becomes the explicit advantage $A_\phi$, and the implicit reward $\hat r_\theta$ becomes the implicit advantage $\hat A_\theta$. GIFT then minimizes the squared difference between these two advantages. Because $z$-score standardization is invariant to additive constants and to multiplicative scalars, the intractable partition function $Z(x)$ (an additive constant within a group at fixed $x$) and the KL coefficient $\beta$ (a multiplicative scalar) both drop out of GIFT's objective. Differentiating the MSE under the on-policy sampling distribution produces a policy-gradient update of the form $\mathbb{E}[w(A_\phi,\hat A_\theta)\,\nabla_\theta\log\pi_\theta(y|x)]$ (Eq.~\ref{eq: GIFT gradient}), placing GIFT in the same algorithmic family as PPO and GRPO while introducing a different advantage weighting.

\paragraph{Contributions.} This paper makes three contributions.
\begin{itemize}
\item \emph{Method.} GIFT is a new on-policy policy-gradient method whose advantage weighting is derived from an MSE between standardized implicit and explicit rewards (Eq.~\ref{eq: GIFT gradient}); it complements rather than replaces PPO and GRPO. Removing the need to tune $\beta$ and to clip gradients, it admits pointwise rather than pairwise supervision, so verifiable rewards, learned reward-model scores, and scalar feedback all plug into the same loss.
\item \emph{Theory.} The population minimizers of $\mathcal{L}_{\text{GIFT}}$ are characterized in closed form (Theorem~\ref{thm:gift-optima}): they coincide with the GRPO/RLHF solution family $\{\pi^{*}_{\beta}\}_{\beta>0}$ with prompt-dependent $\beta(x)=\frac{\sigma_\phi(x)}{\hat{\sigma}_\theta(x)}$. Cancelling $\beta$ in the loss does not abandon the RLHF objective; it transfers $\beta$ from a global hyperparameter to a measurable, variance-driven function $\beta(x)$.
\item \emph{Empirical.} On RLVR (GSM8K, MATH, AIME) at 7B and 32B with matched compute, GIFT converges faster and overfits less than GRPO, DAPO and GSPO. On RLHF (Infinity dataset, AlpacaEval, Arena-Hard) at 7B and 32B, GIFT outperforms GRPO on length-controlled win rates. The two natural token-level realizations of the implicit reward (kl-sum vs.\ kl-average) trade off length sensitivity against length normalization, and kl-sum is adopted throughout.
\end{itemize}

\section{Method}

\subsection{Background}\label{sec:background}

RLHF, RLVR, DPO, UNA, and GRPO all target the same KL-regularized RL objective:
\begin{equation}\label{eq: RL objective}
\pi^{*}(y|x) = \arg\max_{\pi_\theta}\;\mathbb{E}_{x \sim D}\!\left[\mathbb{E}_{y \sim \pi_\theta(y|x)}\!\left(r_\phi(x,y)\right) - \beta\,D_{\text{KL}}\!\left(\pi_\theta(y|x)\,\|\,\pi_{\text{ref}}(y|x)\right)\right]
\end{equation}
whose closed-form optimum at every $\beta>0$ is the reward-tilted reference distribution
\begin{equation}\label{eq: rlhf-path}
\pi^{*}_{\beta}(y|x)=\frac{\pi_{\text{ref}}(y|x)\exp\!\big(\frac{1}{\beta}r_{\phi}(x,y)\big)}{Z_{\beta}(x)}\qquad Z_{\beta}(x)=\sum_{y'}\pi_{\text{ref}}(y'|x)\exp\!\big(\frac{1}{\beta}r_{\phi}(x,y')\big)
\end{equation}
The collection $\{\pi^{*}_{\beta}(\cdot|x):\beta\in(0,\infty)\}$ forms the \emph{GRPO/RLHF solution set} indexed by the inverse KL coefficient $\beta$. PPO~\cite{schulman2017proximalpolicyoptimizationalgorithms} optimizes Eq.~\ref{eq: RL objective} by learning a value model in addition to the policy, reference, and reward; GRPO~\cite{shao2024deepseekmathpushinglimitsmathematical} sidesteps the value model by using the empirical statistics of a group of $N$ on-policy responses as a surrogate for value estimation:
\begin{equation}\label{eq: GRPO}
A_\phi(x,y) = \frac{r_\phi(x,y) - \mu_\phi}{\sigma_\phi}
\end{equation}
where $\mu_\phi,\sigma_\phi$ are the group mean and standard deviation of $r_\phi$. In words: \emph{the explicit advantage is the group $z$-score of the explicit reward}; this is the central operation that converts a raw scalar reward into the variance-normalized signal used by GRPO's policy gradient. DPO~\cite{rafailov2024directpreferenceoptimizationlanguage} reparametrizes the optimum of Eq.~\ref{eq: RL objective} in terms of an implicit reward
\begin{equation}\label{eq: DPO equation}
r_\theta(x,y) = \beta\log\!\frac{\pi_\theta(y|x)}{\pi_{\text{ref}}(y|x)} + \beta\log Z(x)
\end{equation}
in which $\log Z(x)$ is intractable but cancels under pairwise differencing inside the BT likelihood, yielding a pairwise BCE loss that requires preference data. UNA~\cite{wang2025unaunifyingalignmentsrlhfppo} extends DPO to pointwise $(x,y,r)$ data by dropping the $\log Z(x)$ term and using an MSE between $\beta\log(\frac{\pi_\theta(y|x)}{\pi_{\text{ref}}(y|x)})$ and the explicit reward. A complete review of all five methods, with their loss functions, is in Appendix~\ref{app:related}.

\subsection{The GIFT Algorithm}\label{sec:gift-algo}

GIFT combines on-policy group sampling from GRPO, the implicit reward from DPO, and the implicit-vs-explicit MSE from UNA. The unifying construction is symmetric: \emph{the explicit advantage is the $z$-score of the explicit reward}, and \emph{the implicit advantage is the $z$-score of the implicit reward}. The GIFT loss is the squared difference between the two resulting advantages.

For prompt $x$ and response $y$, define the $\beta$-free, $Z$-free implicit reward
\begin{equation}\label{eq: implicit}
\hat r_\theta(x,y) = \log\!\frac{\pi_\theta(y|x)}{\pi_{\text{ref}}(y|x)}
\end{equation}
The sequence-level $\hat r_\theta$ admits two natural token-level realizations. Letting $|y|$ denote the number of tokens in $y$ and $y^{<j}$ the length-$(j{-}1)$ prefix, the \emph{kl-sum} (length-normalization-free) and \emph{kl-average} (length-normalized) variants are
\begin{align}
\hat r_\theta^{\text{sum}}(x,y) &= \sum_{j=1}^{|y|}\log\!\frac{\pi_\theta(y^{j}|x,y^{<j})}{\pi_{\text{ref}}(y^{j}|x,y^{<j})} \label{eq: Implicit Reward Summation}\\
\hat r_\theta^{\text{avg}}(x,y) &= \frac{1}{|y|}\sum_{j=1}^{|y|}\log\!\frac{\pi_\theta(y^{j}|x,y^{<j})}{\pi_{\text{ref}}(y^{j}|x,y^{<j})} \label{eq: Implicit Reward Average}
\end{align}
kl-sum is the sequence-level KL and grows with response length, so it implicitly favors longer responses; kl-average removes the length scaling, making the implicit reward comparable across responses of different lengths but attenuating the divergence signal. kl-sum is adopted throughout because it gives higher pass@1 in the RLVR experiments (Figure~\ref{fig: Hyperparameters}(b)) where the responses can be verified with the ground truth answer.

Let $\hat\mu_\theta,\hat\sigma_\theta$ denote the group mean and standard deviation of $\hat r_\theta$ over the $N$ on-policy samples. In direct parallel with $A_\phi$ in Eq.~\ref{eq: GRPO}, the implicit advantage is the group $z$-score of the implicit reward,
\begin{equation}\label{eq: implicit advantage}
\hat A_\theta(x,y) = \frac{\hat r_\theta(x,y)-sg(\hat\mu_\theta)}{sg(\hat\sigma_\theta)}
\end{equation}
with $sg(\cdot)$ treated as a stop-gradient. The GIFT loss is the population MSE between the two $z$-scored rewards (i.e., between the explicit and implicit advantages):
\begin{equation}\label{eq: GIFT loss}
\mathcal{L}_{\text{GIFT}}(\pi_\theta) = \mathbb{E}_{x\sim D,\,y\sim\pi_\theta(\cdot|x)}\!\left[\big(A_\phi(x,y)-\hat A_\theta(x,y)\big)^{2}\right]
\end{equation}
The training loop is given in Algorithm~\ref{alg:gift}.

\begin{algorithm}[ht]
\caption{GIFT}
\label{alg:gift}
\begin{algorithmic}[1]
\REQUIRE Prompt set $D$, policy $\pi_\theta$, reference $\pi_{\text{ref}}$, reward $r_\phi$ (RLHF) or ground truth $y^*$ (RLVR; $r_\phi(x,y)=\mathbf{1}[y=y^*]$)
\FOR{each $x\in D$}
    \STATE Sample $N$ responses $y_1,\dots,y_N\sim\pi_\theta(\cdot|x)$
    \STATE Compute explicit rewards $r_\phi(x,y_i)$ and implicit rewards $\hat r_\theta(x,y_i)$ via Eq.~\ref{eq: implicit}
    \STATE Group-standardize: $A_\phi(x,y_i)$ via Eq.~\ref{eq: GRPO}, $\hat A_\theta(x,y_i)$ via Eq.~\ref{eq: implicit advantage}
    \STATE Update $\pi_\theta$ to minimize $\mathcal{L}_{\text{GIFT}}$ in Eq.~\ref{eq: GIFT loss}
\ENDFOR
\RETURN $\pi_\theta$
\end{algorithmic}
\end{algorithm}

\paragraph{Gradient: GIFT is a policy-gradient method.} Treating $\hat\mu_\theta,\hat\sigma_\theta$ as stop-gradients, $\nabla_\theta\hat A_\theta(x,y) = \frac{\nabla_\theta\log\pi_\theta(y|x)}{\texttt{sg}(\hat\sigma_\theta)}$. Differentiating Eq.~\ref{eq: GIFT loss} via the product rule yields two contributions: (i) the log-derivative trick applied to the on-policy sampling distribution $y\sim\pi_\theta$, and (ii) the direct gradient through the implicit advantage $\hat A_\theta(x,y)$:
\begin{align}\label{eq: GIFT gradient}
&\nabla_\theta\mathcal{J}_{\text{GIFT}}
= -\nabla_\theta\mathcal{L}_{\text{GIFT}} \nonumber\\
&= -\mathbb{E}_{x\sim D,\,y\sim\pi_\theta(\cdot|x)}\!\left[(A_\phi(x,y)-\hat A_\theta(x,y))^{2}\nabla_\theta\log\pi_\theta(y|x) - \tfrac{2(A_\phi(x,y)-\hat A_\theta(x,y))}{\texttt{sg}(\hat\sigma_\theta)}\nabla_\theta\log\pi_\theta(y|x)\right]\nonumber\\
&= \mathbb{E}_{x\sim D,\,y\sim\pi_\theta(\cdot|x)}\!\left[\left(\tfrac{2(A_\phi(x,y)-\hat A_\theta(x,y))}{\texttt{sg}(\hat\sigma_\theta)} - (A_\phi(x,y)-\hat A_\theta(x,y))^{2}\right)\nabla_\theta\log\pi_\theta(y|x)\right]
\end{align}

Equation~\ref{eq: GIFT gradient} has the canonical \emph{score-function} (REINFORCE) structure $\mathbb{E}[w(x,y)\,\nabla_\theta\log\pi_\theta(y|x)]$ used by PPO and GRPO. GIFT is therefore a policy-gradient method--it does not replace policy-gradient methods, it adds a new advantage weighting $w(x,y)=\frac{2(A_\phi(x,y)-\hat A_\theta(x,y))}{sg(\hat\sigma_\theta)} - (A_\phi(x,y)-\hat A_\theta(x,y))^{2}$ derived from the MSE in Eq.~\ref{eq: GIFT loss}, in contrast to GRPO's $A_\phi(x,y)$ alone. Because $A_\phi(x,y)$ and $\hat A_\theta(x,y)$ are both $z$-scored rewards, the weight $w$ is fully expressed in terms of advantages--no raw rewards or hyperparameters appear.

\begin{proposition}[$\beta$ and $Z(x)$ cancel under $z$-score standardization]\label{lem:cancellation}
The group $z$-score of the DPO implicit reward $r_\theta(x,y)=\beta\log\frac{\pi_\theta(y|x)}{\pi_{\text{ref}}(y|x)}+\beta\log Z(x)$ equals the group $z$-score of the $\beta$-free implicit reward $\hat{r}_\theta(x,y)=\log\frac{\pi_\theta(y|x)}{\pi_{\text{ref}}(y|x)}$, so the implicit advantage $\hat{A}_\theta(x,y)=A_\theta(x,y)$ is independent of both $\beta$ and $Z(x)$.
\end{proposition}
Within a fixed group at prompt $x$, the term $\beta\log Z(x)$ is constant across all responses and vanishes under mean subtraction (Eq.~\ref{eq: DPO_mean}-\ref{eq: GIFT_mean_subtraction}); the remaining factor $\beta$ scales both the centered reward and its standard deviation identically, so it cancels in the ratio $(\text{centered})/(\text{std})$ (Eq.~\ref{eq: GIFT_variance}-\ref{eq: GIFT Implicit Reward Normalization}). Formally, $z$-score standardization maps any $cX+d$ to the same $z$-score as $X$, since $\frac{(cX+d)-\mathbb{E}[cX+d]}{\sqrt{\mathrm{Var}(cX+d)}}=\frac{X-\mathbb{E}[X]}{\sqrt{\mathrm{Var}(X)}}$ for any $c>0,\,d\in\mathbb{R}$. The full algebra is in Appendix~\ref{app:proof-cancellation}.

\subsection{What GIFT Optimizes: A Variance-Determined \texorpdfstring{$\beta(x)$}{β(x)} on the GRPO Path}\label{sec:opt-target}

A natural objective is that, because $\mathcal{L}_{\text{GIFT}}$ is independent of any $\beta$ scalar, it cannot be solving the $\beta$-regularized objective in Eq.~\ref{eq: RL objective}. Theorem~\ref{thm:gift-optima} resolves this: matching the two $z$-scored rewards (the implicit and explicit advantages) forces the policy onto the GRPO/RLHF solution set $\{\pi^{*}_{\beta}\}_{\beta>0}$, but with a prompt-dependent $\beta(x)$ determined endogenously by the variance ratio of the rewards before $z$-scoring. Let $\mathcal{Y}(x)$ denote the (finite) response set for $x$, and let $\sigma_\phi(x)$ and $\hat\sigma_\theta(x)$ denote the standard deviations of the explicit reward $r_\phi(x,\cdot)$ and the implicit reward $\hat r_\theta(x,\cdot)$ under $y\sim\pi_\theta(\cdot|x)$.

\begin{assumption}[Non-degeneracy]\label{ass:nondeg}
For every prompt $x$ in the prompt dataset $D$: the explicit reward $r_{\phi}(x,\cdot)$ is non-constant on $\mathcal{Y}(x)$, and $\sigma_\phi(x)>0$.
\end{assumption}

This assumption is typically enforced through dynamic filtering \cite{yu2025dapoopensourcellmreinforcement}, where prompts whose sampled responses receive identical rewards—and therefore yield zero advantage and zero variance—are removed from training.

\begin{theorem}[GIFT optima coincide with the GRPO path; $\beta(x)$ is variance-determined]\label{thm:gift-optima}
Under Assumption~\ref{ass:nondeg}: \textup{(a)} for every $\beta>0$, $\pi^{*}_{\beta}$ achieves $A_{\phi}(x,y)=\hat A_{\theta}(x,y)$ pointwise, so $\mathcal{L}_{\text{GIFT}}(\pi^{*}_{\beta})=0$; \textup{(b)} if $A_{\phi}(x,y)=\hat A_{\theta}(x,y)$ for all $y\in\mathcal{Y}(x)$, then $\pi_\theta(\cdot|x)=\pi^{*}_{\beta(x)}(\cdot|x)$ with
\begin{equation}\label{eq: beta-x}
\beta(x)\;=\;\frac{\sigma_\phi(x)}{\hat{\sigma}_\theta(x)}\;\in\;(0,\infty)
\end{equation}
\textup{(c)} $\mathcal{L}_{\text{GIFT}}(\pi_\theta)\ge 0$, with equality iff $\pi_\theta(\cdot|x)=\pi^{*}_{\beta(x)}(\cdot|x)$ for some $\beta(x)>0$ and $D$-almost every $x$.
\end{theorem}
The proof rests on $z$-score equality being equivalent to an affine relation, and is in Appendix~\ref{app:proof-thm-optima}. Theorem~\ref{thm:gift-optima} shows GIFT and GRPO target the same parametric policy family, but where GRPO fixes a global hyperparameter $\beta$, GIFT determines $\beta(x)=\frac{\sigma_\phi(x)}{\hat{\sigma}_\theta(x)}$ endogenously by matching the dispersion of the explicit and implicit rewards under the converged on-policy distribution. Cancelling $\beta$ from the loss therefore does not eliminate $\beta$ from the solution--it transfers $\beta$ from a hyperparameter into a measurable, prompt-adaptive, variance-driven function of the converged policy.

\subsection{Comparison with PPO, GRPO, DPO, and UNA}\label{sec:comparison}

\begin{figure}[!hbtp]
    \centering
    \includegraphics[width=0.8\linewidth]{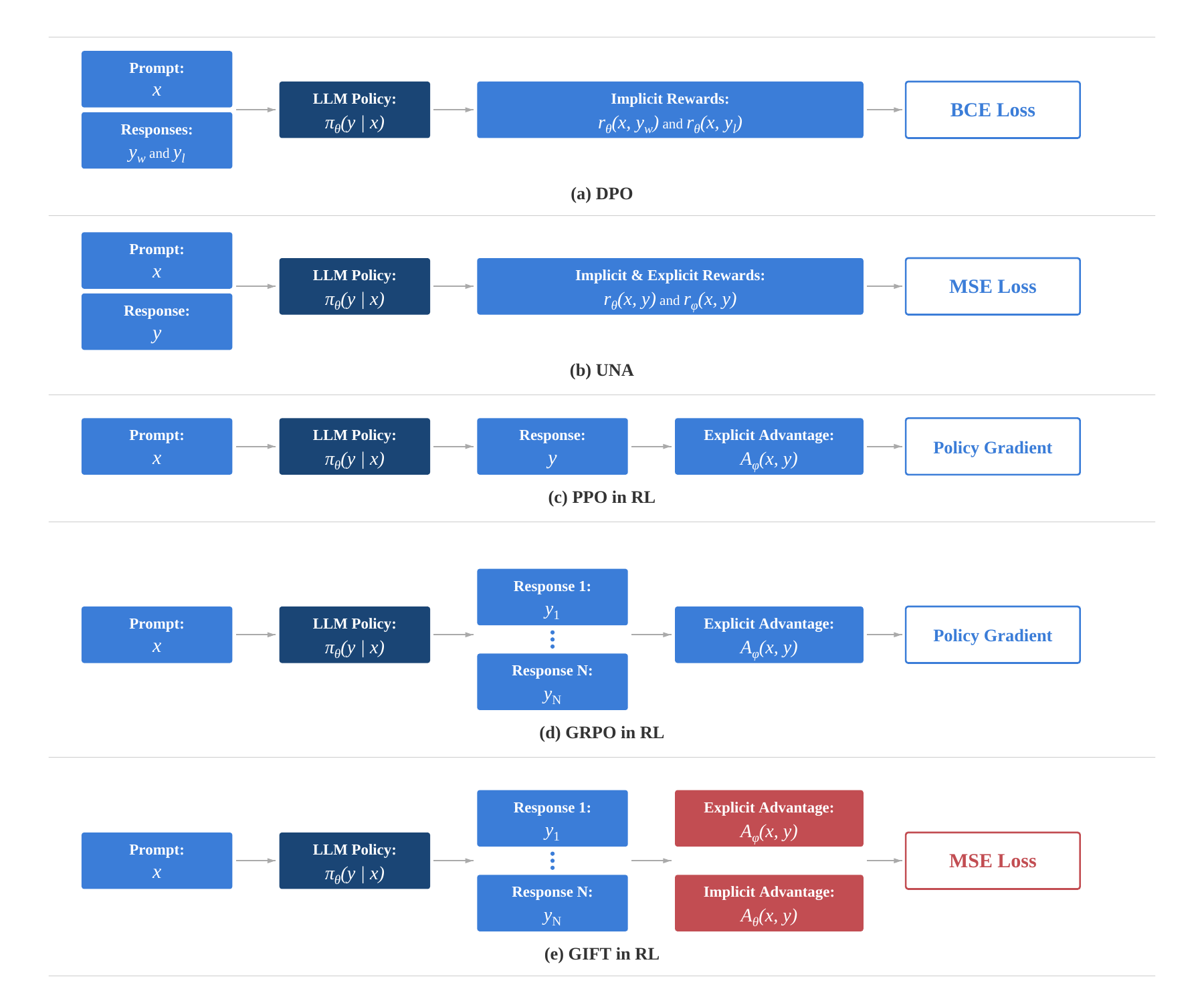}
    \caption{Conceptual comparison of (a) DPO, (b) UNA, (c) PPO, (d) GRPO, and (e) GIFT.}
    \label{fig: GIFT}
\end{figure}

GIFT borrows on-policy group sampling from GRPO, the implicit reward from DPO, and the implicit-vs-explicit MSE from UNA. PPO, GRPO, and GIFT are all on-policy policy-gradient methods; they differ in (a) what is used as the advantage and (b) how the advantage weights $\nabla_\theta\log\pi_\theta(y|x)$. PPO uses a learned value-model baseline to convert the explicit reward into an advantage; GRPO replaces the value model with a $z$-score, so its explicit advantage $A_\phi(x,y)$ is the group $z$-score of $r_\phi(x,y)$; GIFT goes one step further and applies the \emph{same} $z$-score map to the implicit reward $\hat r_\theta(x,y)$, producing the implicit advantage $\hat A_\theta(x,y)$, and then matches the two advantages via MSE. Figure~\ref{fig: GIFT} visualizes the conceptual differences across the five methods, and Table~\ref{tab:method_comparison} summarizes their structural properties. Practically, compared to GRPO:
\begin{itemize}
\item \emph{Loss surrogate.} GIFT and GRPO both produce score-function policy gradients; GIFT's advantage weighting is $\frac{2(A_\phi(x,y)-\hat A_\theta(x,y))}{sg(\hat\sigma_\theta)} - (A_\phi(x,y)-\hat A_\theta(x,y))^2$ from Eq.~\ref{eq: GIFT gradient}, while GRPO's is the clipped explicit advantage $A_\phi(x,y)$. GIFT requires no clipping in the experiments.
\item \emph{Hyperparameters.} GIFT has no $\beta$ to tune (Proposition~\ref{lem:cancellation}) and no clipping parameter, while GRPO selects a fixed KL coefficient $\beta$ in advance.
\item \emph{Convergence and generalization.} Empirically GIFT converges faster and shows a smaller train-eval gap on GSM8K/MATH/AIME (Section~\ref{sec:experiments}).
\end{itemize}

GIFT differs fundamentally from prior $\beta$-free preference optimization methods such as SimPO \cite{meng2024simposimplepreferenceoptimization} and ORPO \cite{hong2024orpomonolithicpreferenceoptimization}. These methods operate in an offline pairwise setting and obtain $\beta$-free objectives through, respectively, length-normalized preference margins and odds-ratio regularization added to SFT. 
Among on-policy approaches, RAFT \cite{dong2023raftrewardrankedfinetuning} employs rejection sampling over on-policy generations, while several subsequent methods \cite{minimax2025minimaxm1scalingtesttimecompute, zhao2025geometricmeanpolicyoptimization, roux2025taperedoffpolicyreinforcestable} explicitly remove KL regularization constraints. In contrast, GIFT introduces an MSE-derived policy gradient objective and performs group-wise $z$-score standardization over both implicit and explicit rewards.

\begin{table}[ht]
\centering
\small
\caption{Structural comparison of optimization methods. ``$\beta$-free loss'' refers to algebraic independence of the loss from the KL coefficient $\beta$.}
\label{tab:method_comparison}
\begin{tabular}{l|cccccc}
\toprule
\textbf{Method} & \textbf{On Policy} & \textbf{Policy Grad.} & \textbf{Group Norm.} & \textbf{Data Type} & \textbf{$\beta$-free} & \textbf{Loss surrogate} \\
\midrule
DPO  & No  & No  & No  & Pairwise  & No  & BCE \\
UNA  & No  & No  & No  & Pointwise & No  & MSE \\
PPO  & Yes & Yes & No  & Pointwise & No  & reward maximization \\
GRPO & Yes & Yes & Yes & Pointwise & No  & reward maximization \\
GIFT & Yes & Yes & Yes & Pointwise & Yes & MSE \\
\bottomrule
\end{tabular}
\end{table}

\section{Experiments}\label{sec:experiments}

\begin{figure}[!hbtp]
    \centering
    \includegraphics[width=0.8\linewidth]{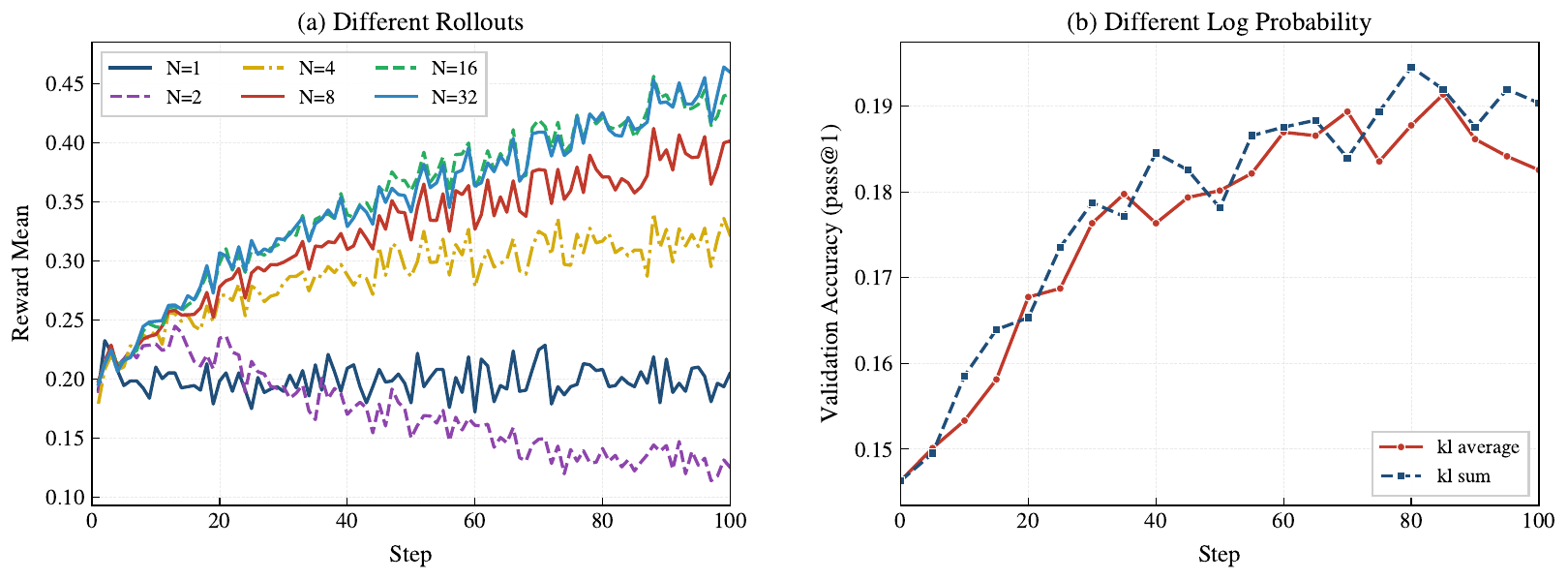}
    \caption{(a) Effect of rollout number $N\in\{1,2,4,8,16,32\}$. (b) kl-sum vs.\ kl-average implicit reward with standard-deviation normalization.}
    \label{fig: Hyperparameters}
\end{figure}

\paragraph{Setup.} GRPO, DAPO and GSPO are the appropriate baseline because it shares the same on-policy training protocol, data pipeline, and per-step compute as GIFT. RLVR experiments use \texttt{deepseek-llm-7b-chat}~\cite{deepseekai2025deepseekr1incentivizingreasoningcapability} and \texttt{Qwen2.5-32B-Instruct}~\cite{qwen2.5} on GSM8K~\cite{cobbe2021gsm8k} and MATH-lighteval~\cite{hendrycksmath2021}.
Then, Qwen3-30BA3B-base ~\cite{yang2025qwen3technicalreport} and Qwen3.5-35BA3B-base \cite{qwen3.5} are trained on DAPO's math dataset ~\cite{yu2025dapoopensourcellmreinforcement} and evaluation on AIME 2024 ~\cite{aime2024_huggingface}, AIME 2025 ~\cite{aime25} and AMC 2023 ~\cite{math-ai_amc23}. RLHF uses the Infinity dataset~\cite{li2025infinityinstructscalinginstruction} on \texttt{Qwen2.5-7B/32B-Instruct}~\cite{qwen2.5} with the Skywork reward model~\cite{liu2025skywork}. More detailed hyperparameter settings can be found in Appendix \ref{app:training-config}.

\subsection{Variants of GIFT}\label{subsec:variants_GIFT}

Sweeping $N\in\{1,2,4,8,16,32\}$, group statistics become more accurate as $N$ grows; Figure~\ref{fig: Hyperparameters}(a) shows pass@1 improving with diminishing gains beyond $N=16$, which is adopted henceforth. Comparing the kl-sum and kl-average implicit-reward variants (Figure~\ref{fig: Hyperparameters}(b)), kl-sum gives higher pass@1 and is adopted.

\begin{figure}[!hbtp]
    \centering
    \includegraphics[width=0.8\linewidth]{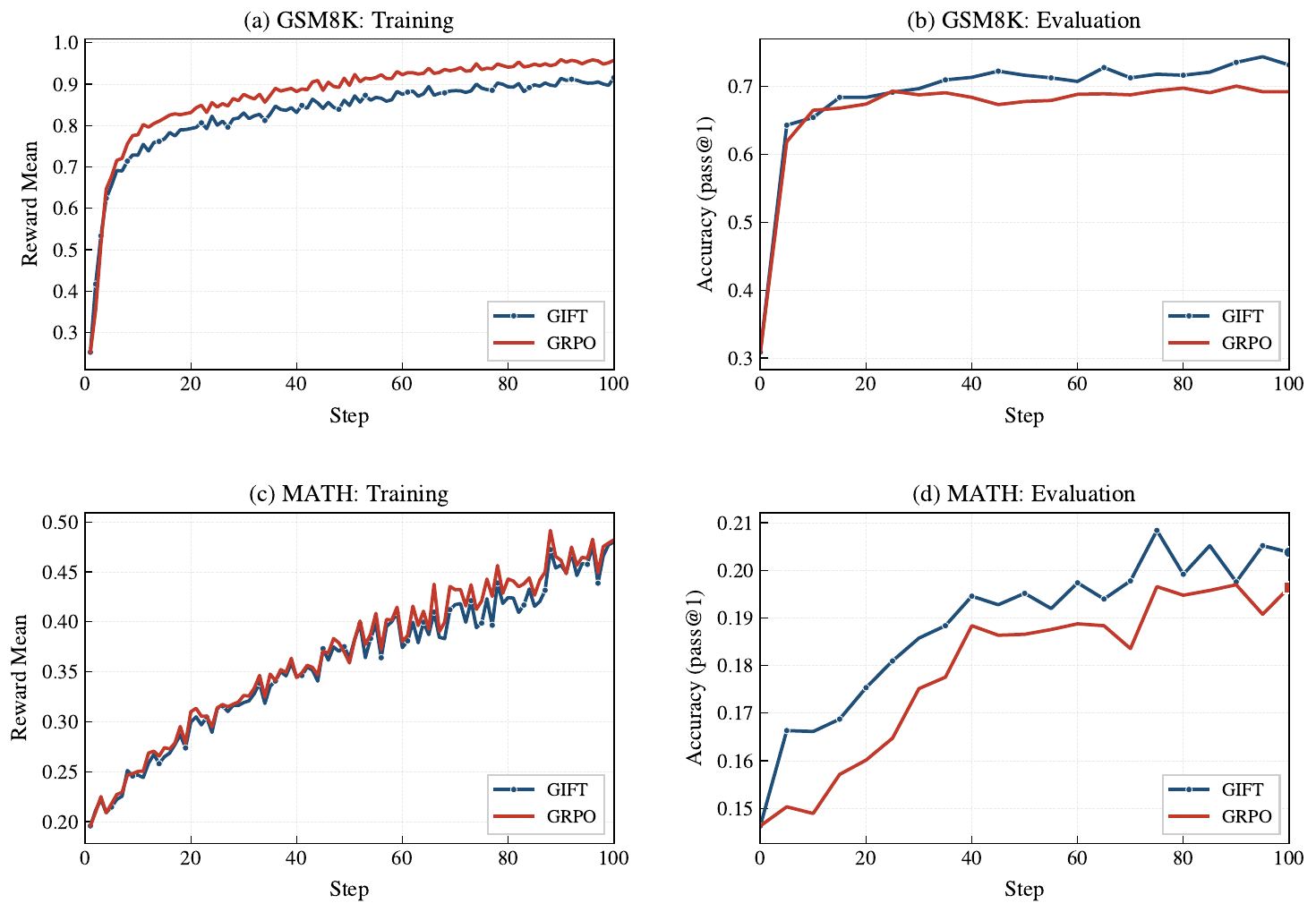}
    \caption{GIFT vs.\ GRPO on DeepSeek-7B (GSM8K, MATH). GRPO shows stronger overfitting: higher training accuracy but lower evaluation accuracy on GSM8K.}
    \label{fig: 7B}
\end{figure}

\begin{figure}[!hbtp]
    \centering
    \includegraphics[width=0.8\linewidth]{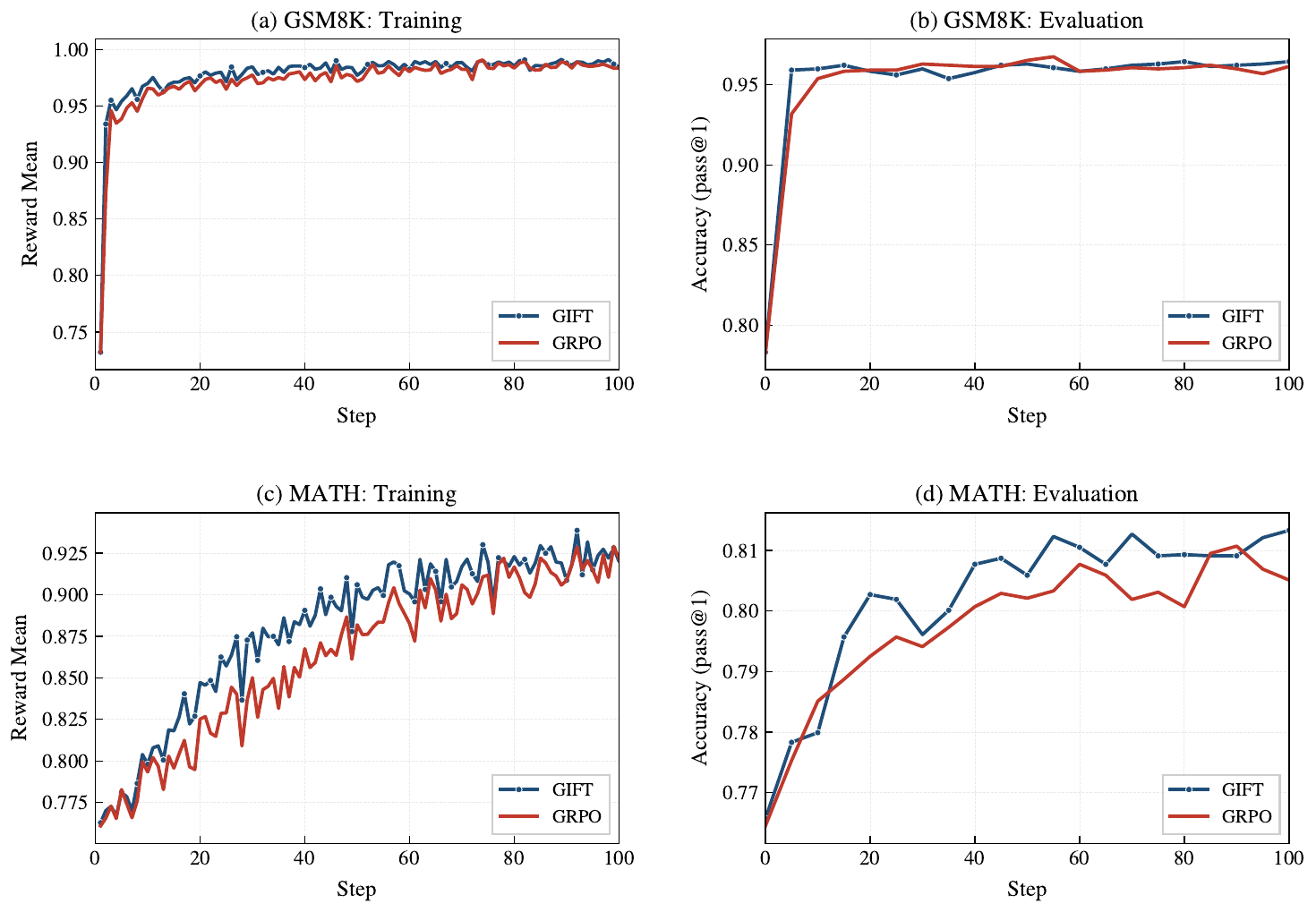}
    \caption{GIFT vs.\ GRPO on Qwen2.5-32B (GSM8K, MATH). GIFT achieves faster convergence and a smaller train-eval gap.}
    \label{fig: 32B}
\end{figure}

\subsection{RLVR: GIFT vs.\ GRPO on GSM8K and MATH}\label{sec:rlvr-results-GSM8K-MATH}

Across DeepSeek-7B and Qwen2.5-32B on GSM8K and MATH (training/evaluation curves in Figures~\ref{fig: 7B} and~\ref{fig: 32B}), GIFT converges faster and shows a smaller train-eval gap than GRPO; on GSM8K with DeepSeek-7B, GRPO has higher training accuracy but lower evaluation accuracy, consistent with stronger overfitting.

\begin{figure}[!hbtp]
    \centering
    \includegraphics[width=0.9\linewidth]{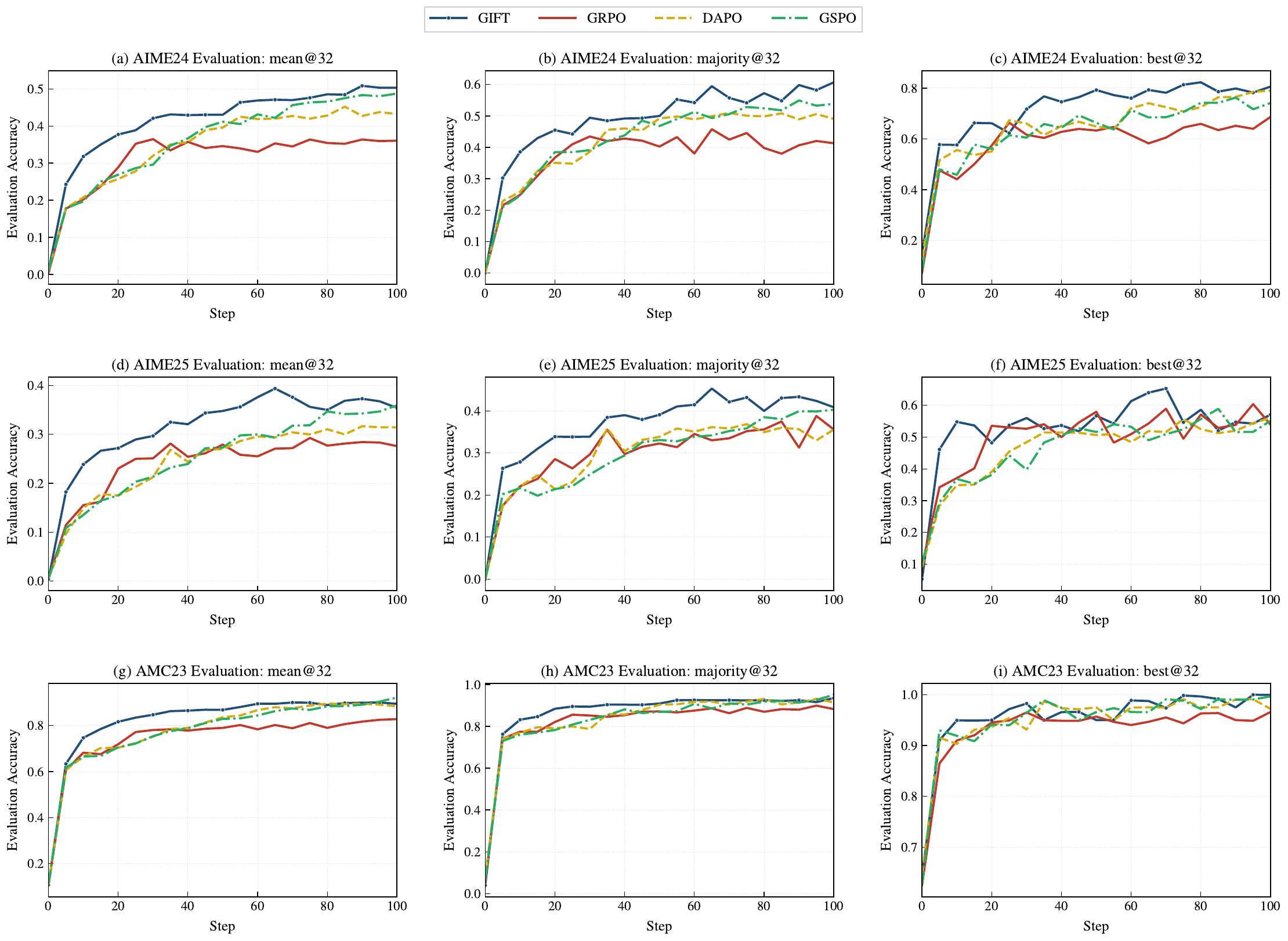}
    \caption{GIFT vs.\ GRPO, DAPO and GSPO trained on DAPO math training dataset and evaluated on AIME2024, AIME2025 and AMC2023 with base Qwen3 30BA3B model.}
    \label{fig:AIME-Qwen3}
\end{figure}

\begin{figure}[!hbtp]
    \centering
    \includegraphics[width=0.9\linewidth]{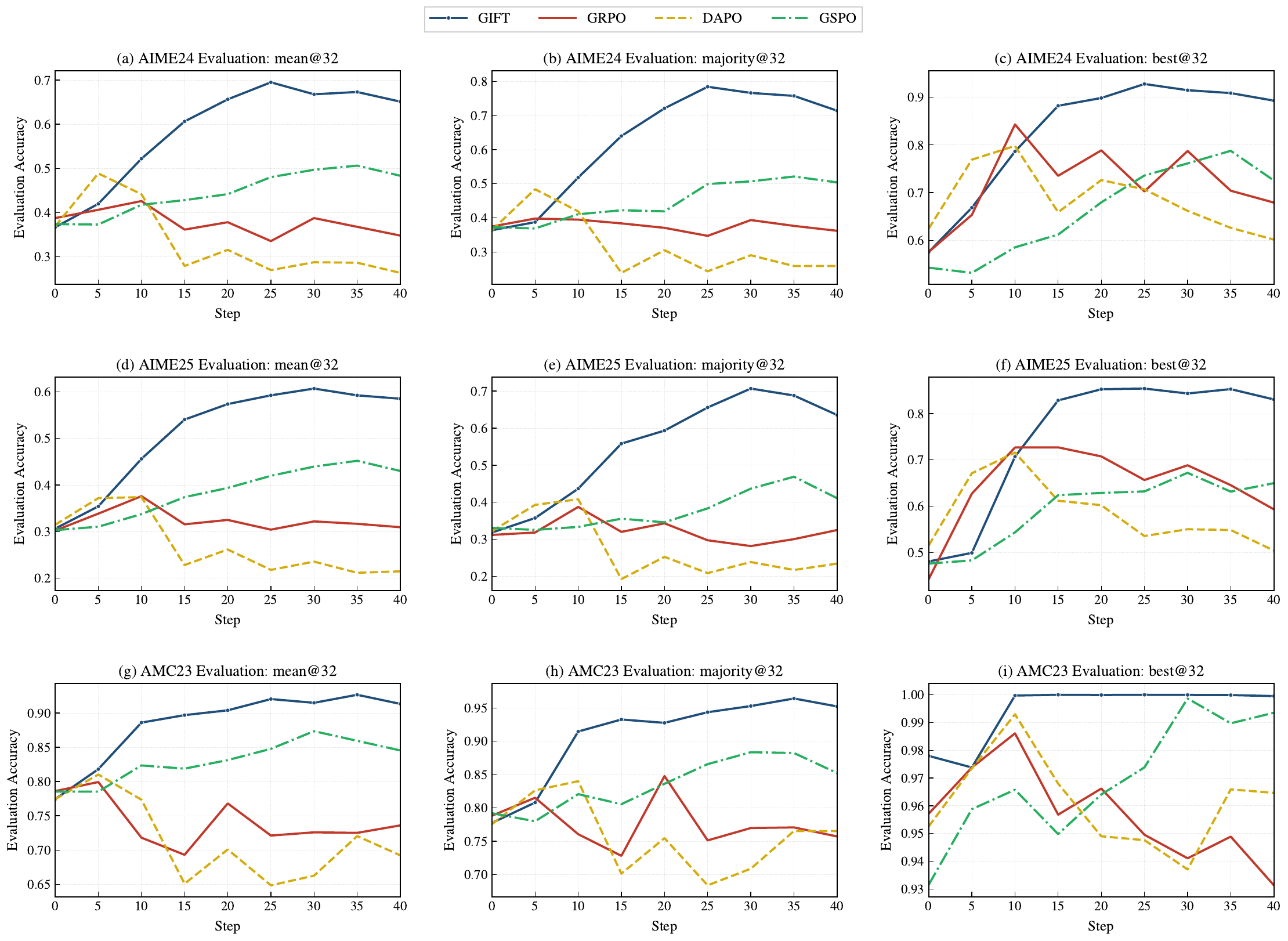}
    \caption{GIFT vs.\ GRPO, DAPO and GSPO trained on DAPO math training dataset and evaluated on AIME2024, AIME2025 and AMC2023 with base Qwen3.5 35BA3B model.}
    \label{fig:AIME-Qwen3.5}
\end{figure}

\subsection{RLVR: GIFT vs.\ GRPO, DAPO and GSPO on AMIE and AMC}\label{sec:rlvr-results-AIME-AMC}

For more challenging mathematical reasoning tasks, GIFT is compared against GRPO, DAPO, and GSPO. Performance is evaluated using three metrics: \texttt{mean@32}, which measures the average response accuracy across 32 rollouts; \texttt{maj@32}, which measures the majority-voting accuracy; and \texttt{best@32}, which measures the best response accuracy among the 32 sampled rollouts. As shown in Figure~\ref{fig:AIME-Qwen3} and Figure~\ref{fig:AIME-Qwen3.5}, GIFT consistently achieves faster convergence and stronger overall performance than GRPO, DAPO, and GSPO. Moreover, the performance gains are more pronounced on Qwen3.5 than on Qwen3.

\subsection{RLHF: GIFT vs.\ GRPO}\label{sec:rlhf-results}

\begin{figure}[!hbtp]
    \centering
    \includegraphics[width=0.9\linewidth]{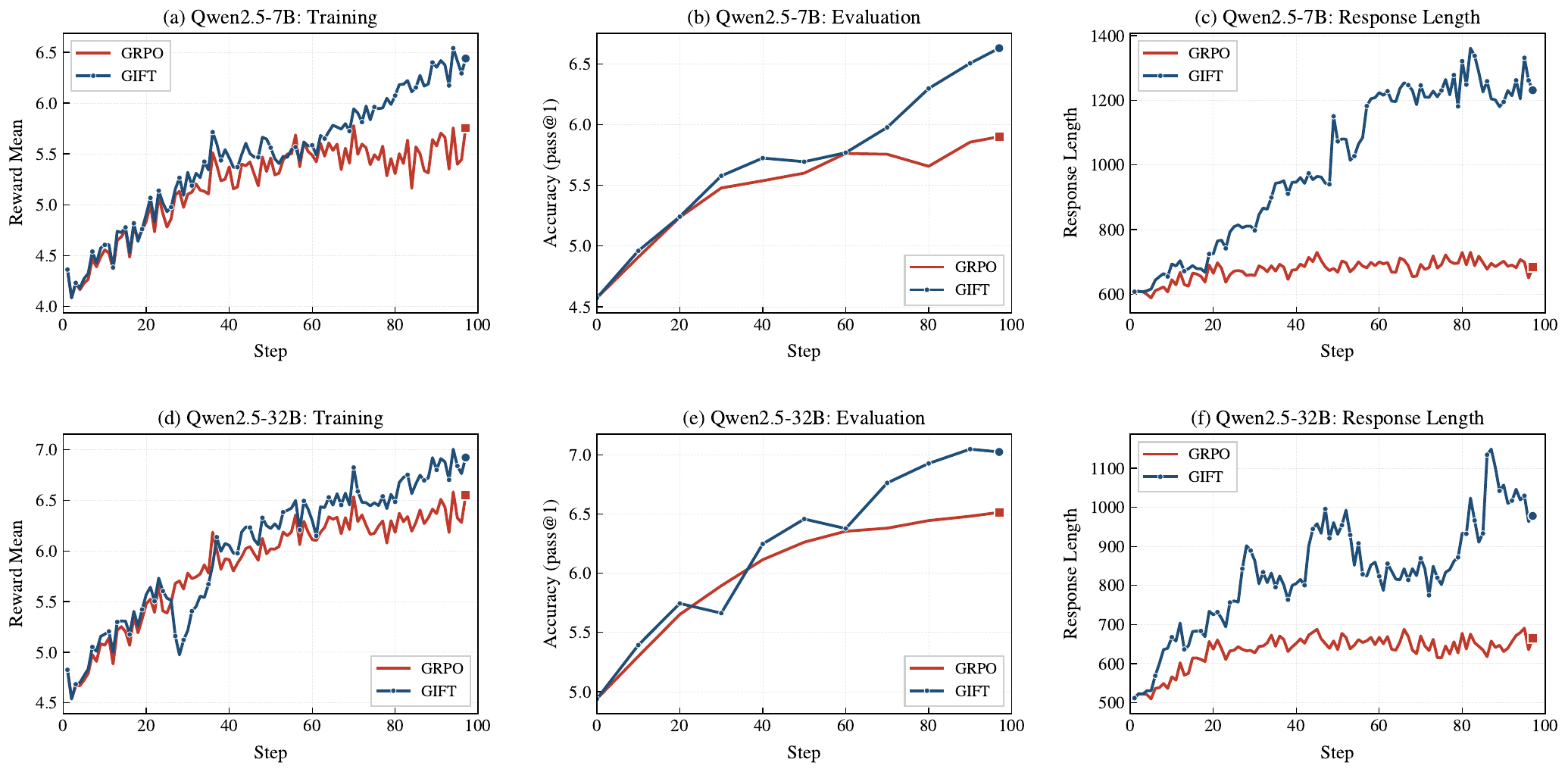}
    \caption{GIFT vs.\ GRPO under RLHF on Qwen2.5-7B-Instruct and Qwen2.5-32B-Instruct using the Infinity dataset.}
    \label{fig: RLHF_GIFT}
\end{figure}

Beyond RLVR, GIFT is applied to RLHF for instruction-following improvements. Learned reward models exhibit length bias, so the explicit reward is normalized by $\sqrt{|y|}$; the implicit reward remains kl-sum. Figure~\ref{fig: RLHF_GIFT} shows GIFT outperforming GRPO on Qwen2.5-7B and Qwen2.5-32B during training. Tables~\ref{tab:gift_selected_results} and~\ref{tab:model_performance} report downstream evaluations: GIFT matches or improves over GRPO on all seven of TruthfulQA~\cite{lin2022truthfulqameasuringmodelsmimic}, BBQ~\cite{parrish-etal-2022-bbq}, MBPP~\cite{austin2021programsynthesislargelanguage}, ARC-Challenge~\cite{clark2018thinksolvedquestionanswering}, Winogender~\cite{rudinger2018genderbiascoreferenceresolution}, GPQA~\cite{rein2023gpqagraduatelevelgoogleproofqa}, and MUSR~\cite{sprague2024musrtestinglimitschainofthought}, and beats GRPO on AlpacaEval~\cite{dubois2024length} length-controlled win rate (48.77 vs.\ 35.33 at 7B; 61.43 vs.\ 53.13 at 32B) and Arena-Hard~\cite{arenahard2024} style-control at 32B (15.3 vs.\ 7.7).

\paragraph{Length-controlled gains.} GIFT produces longer responses than GRPO under RLHF (4291 vs.\ 2203 tokens at 7B; 3327 vs.\ 2369 at 32B). The length-controlled metrics, designed precisely to neutralize length effects, confirm the quality gain: GIFT outperforms GRPO on the AlpacaEval length-controlled win rate at both 7B and 32B and on Arena-Hard style-control at 32B (15.3 vs.\ 7.7).

\begin{table}[ht]
\centering
\small
\caption{Downstream evaluations across seven benchmarks. Bold indicates the best within a model size. Several gaps are within single-seed noise.}
\label{tab:gift_selected_results}
\begin{tabular}{lccccccc}
\hline
\textbf{Model}
& \textbf{TruthfulQA}
& \textbf{BBQ}
& \textbf{MBPP}
& \textbf{ARC-C}
& \textbf{Winogender}
& \textbf{GPQA}
& \textbf{MUSR} \\
\hline
7B-Instruct  & 64.79 & 54.62 & 48.2 & 64.59 & 60.97 & 31.46 & 42.86 \\
7B-GRPO      & 66.93 & 53.79 & 62.0 & 64.25 & 59.86 & 30.87 & 45.77 \\
7B-GIFT      & \textbf{69.05} & \textbf{58.08} & \textbf{62.8} & \textbf{65.7} & \textbf{61.39} & \textbf{33.98} & \textbf{46.16} \\
\hline
32B-Instruct & 65.58 & 59.04 & 75.8 & 71.93 & 61.11 & \textbf{38.26} & \textbf{50.13} \\
32B-GRPO     & 69.09 & 65.69 & 75.8 & 72.27 & 64.31 & 37.16 & 48.68 \\
32B-GIFT     & \textbf{70.02} & \textbf{71.06} & \textbf{78.2} & \textbf{73.89} & \textbf{64.72} & 37.25 & 49.47 \\
\hline
\end{tabular}
\end{table}

\begin{table}[ht]
\centering
\small
\caption{AlpacaEval and Arena-Hard. LC = length-controlled.}
\label{tab:model_performance}
\begin{tabular}{lccccc}
\hline
\textbf{Model}
& \multicolumn{2}{c}{\textbf{AlpacaEval}}
& \multicolumn{3}{c}{\textbf{Arena-Hard}} \\
\cline{2-3}\cline{4-6}
& \textbf{LC Win Rate} & \textbf{Win Rate} & \textbf{Resp.\ Length} & \textbf{Style Ctrl.} & \textbf{Creative Writing} \\
\hline
7B-Instruct  & 30.81 & 29.05 & 1989 & 2.5 & 5.2 \\
7B-GRPO      & 35.33 & 38.25 & 2203 & \textbf{3.2} & 8.5 \\
7B-GIFT      & \textbf{48.77} & \textbf{72.43} & 4291 & 3.0 & \textbf{17.2} \\
\hline
32B-Instruct & 43.49 & 35.76 & 1754 & 4.4 & 3.9 \\
32B-GRPO     & 53.13 & 60.77 & 2369 & 7.7 & 21.1 \\
32B-GIFT     & \textbf{61.43} & \textbf{78.53} & 3327 & \textbf{15.3} & \textbf{47.9} \\
\hline
\end{tabular}
\end{table}

\section{Conclusion}\label{sec:conclusion}

GIFT applies group $z$-score standardization to both the explicit reward $r_\phi(x,y)$ and the DPO-style implicit reward $\hat r_\theta(x, y)=\log(\frac{\pi_\theta(y|x)}{\pi_{\text{ref}}(y|x)})$-turning each into an advantage--and minimizes the squared difference between the resulting implicit and explicit advantages, with $Z(x)$ and $\beta$ dropping out of the loss as a mechanical property of $z$-score standardization. Differentiating this loss under the on-policy sampling distribution produces a score-function policy gradient (Eq.~\ref{eq: GIFT gradient}), so GIFT is a policy-gradient method in the same family as PPO and GRPO; it complements rather than replaces them, and differs only in the advantage weighting placed on $\nabla_\theta\log\pi_\theta(y|x)$. Under matched compute, GIFT converges faster than GRPO, DAPO and GSPO, has a smaller train-eval gap, and produces stronger length-controlled win rates under RLHF, without a $\beta$ or clipping hyperparameter. Theorem~\ref{thm:gift-optima} shows GIFT solves the same GRPO/RLHF policy family $\{\pi^{*}_{\beta}\}_{\beta>0}$ with a prompt-dependent $\beta(x)=\frac{\sigma_\phi(x)}{\hat{\sigma}_\theta(x)}$ determined endogenously by reward-variance matching--it does not abandon the RLHF objective; it replaces a global hyperparameter with a variance-driven function $\beta(x)$.

\paragraph{Limitations.} Theorem~\ref{thm:gift-optima} characterizes the set of minimizers of $\mathcal{L}_{\text{GIFT}}$, but does not uniquely determine which $\pi^{*}_{\beta(x)}$ along the GRPO/RLHF solution path is selected at convergence. In addition, the theoretical properties of the KL-sum and KL-average formulations remain an interesting direction for further analysis. Future work may also explore scaling behavior under larger model sizes, longer prompts and responses, and broader training and evaluation datasets.

{
\small
\bibliographystyle{unsrtnat}
\bibliography{references}
}


\appendix

\section{Background Methods in Detail}\label{app:related}

\subsection{RLHF and PPO}
After the pretraining and SFT stages, LLMs may still produce undesirable or suboptimal responses. To further improve their response quality, RLHF is applied. RLHF typically consists of two major stages, each leveraging a distinct dataset. In the reward model training stage, the dataset comprises triplets of the form $(x, y_w, y_l)$, where $x$ denotes the prompt, $y_w$ represents the preferred (winning) response, and $y_l$ represents the dispreferred (losing) response. In the RL fine-tuning stage, the dataset contains only task prompts $x$, which correspond to the domains or behaviors that the LLM aims to improve through policy optimization.

During reward model training, an explicit reward model is learned to predict a scalar score $r_{\phi}(x, y)$ for a given prompt-response pair $(x, y)$. The model is trained on pairwise preference data $(x, y_w, y_l)$, where $y_w$ denotes the preferred response and $y_l$ the dispreferred one. The probability that $y_w$ is favored over $y_l$, denoted as $P_{\phi}(y_w > y_l \mid x)$, is modeled using the Bradley-Terry (BT) framework, i.e., $P_{\phi}(y_w > y_l \mid x) = \frac{e^{r_{\phi}(x, y_w)}}{e^{r_{\phi}(x, y_w)} + e^{r_{\phi}(x, y_l)}} = \sigma\!\left(r_{\phi}(x, y_w) - r_{\phi}(x, y_l)\right)$ where $\sigma(\cdot)$ denotes the logistic sigmoid function~\cite{bradley1952rank}. Intuitively, the higher the difference $r_{\phi}(x, y_w) - r_{\phi}(x, y_l)$, the greater the likelihood that the model predicts $y_w$ as more aligned with human preference. Eventually, the RM is trained through a binary cross entropy (BCE) loss $L_{\text{RM}}(\pi_\theta) = - \mathbb{E}_{(x, y_w, y_l) \sim D}\left[ \log \sigma\!\left(r_{\phi}(x, y_w) - r_{\phi}(x, y_l)\right) \right]$~\cite{ouyang2022traininglanguagemodelsfollow}.

RL fine-tuning serves two principal objectives. First, it seeks to maximize the pretrained explicit reward function $r_{\phi}(x, y)$ so that the policy $\pi_\theta(y|x)$ becomes better aligned with human preferences encoded by the reward model. Second, to mitigate \emph{reward hacking} and preserve fidelity to the pretrained model, a Kullback-Leibler (KL) regularization term is introduced, penalizing large deviations from the reference policy $\pi_{\text{ref}}(y|x)$. The combined optimization problem can be expressed as
\begin{equation}
\label{eq: appendix RL objective}
\pi^*(y|x) = \arg\max_{\pi_\theta} \mathbb{E}_{x \sim D} \left[ \mathbb{E}_{y \sim \pi_\theta(y|x)} \left( r_\phi(x, y) \right) - \beta D_{\text{KL}}\!\left(\pi_\theta(y|x) \,\|\, \pi_{\text{ref}}(y|x)\right) \right]
\end{equation}
where $\beta$ is a KL coefficient that controls the trade-off between reward maximization and policy divergence. \textbf{The same objective is adopted in RLVR, DPO, UNA and GIFT.}

To stabilize training, Proximal Policy Optimization (PPO)~\cite{schulman2017proximalpolicyoptimizationalgorithms} is commonly adopted to optimize the objective in Equation~\ref{eq: appendix RL objective}. However, RLHF requires maintaining multiple large-scale components--namely, a policy model, reference model, reward model, and value model--resulting in substantial computational and memory overhead. These drawbacks significantly restrict the scalability and practicality of RLHF for LLMs.

\subsection{RLVR and GRPO}
While RLHF has proven effective at aligning LLMs with human preferences and mitigating undesirable or harmful outputs, it often falls short in enhancing complex reasoning capabilities such as mathematical problem solving or program synthesis. To overcome this limitation, RLVR has been proposed, which only has the RL fine-tuning stage as the explicit reward model has been replaced with verifiable reward. In the RL fine-tuning stage of RLVR, the dataset contains $(x, y^*)$ where $y^*$ refers to the golden verifiable response. In contrast to RLHF, which depends on a learned and potentially imperfect reward model trained from human feedback, RLVR leverages verifiable signals--such as the correctness of a numerical solution or the success of generated code in passing predefined test cases--as rewards $r_\phi(x, y, y^*) =
\begin{cases}
1, & \text{if } y = y^* \\
0, & \text{if } y \ne y^*
\end{cases}$. These verifiable signals, although sparse, provide precise and unambiguous feedback that directly reflects task success.

To reduce the computation cost of PPO, GRPO is then proposed~\cite{shao2024deepseekmathpushinglimitsmathematical, zheng2025groupsequencepolicyoptimization} as shown in Equation~\ref{eq: appendix GRPO}, where $A_\phi(x, y)$ refers to the normalized explicit reward, $\mu_\phi$ refers to the mean of the explicit reward and $\sigma_\phi$ refers to the standard deviation of the explicit reward across an on-policy group of responses.
\begin{equation}
\label{eq: appendix GRPO}
A_\phi(x, y) = \frac{r_\phi(x,y) - \mu_\phi}{\sigma_\phi}
\end{equation}
When combined with the GRPO framework, as demonstrated in DeepSeek R1~\cite{deepseekai2025deepseekr1incentivizingreasoningcapability}, RLVR substantially enhances an LLM's reasoning capacity. This integration enables the model to sustain longer and more coherent chains of thought (CoT), exhibit improved self-consistency across reasoning steps, and even display emergent ``aha moments'' where the model identifies and corrects its own logical or computational errors during inference.

\subsection{DPO}
To address the instability and inefficiency of RLHF and PPO, DPO was proposed as a simplified offline alternative utilizing the same reward model training dataset $(x, y_w, y_l)$ from RLHF~\cite{rafailov2024directpreferenceoptimizationlanguage}. Based on the RL objective in Equation~\ref{eq: appendix RL objective}, DPO eliminates the explicit reward model and builds a mapping between an implicit reward model and the optimal policy as shown in Equation~\ref{eq: appendix DPO equation}:
\begin{equation}
\label{eq: appendix DPO equation}
r_\theta(x, y) = \beta \log\!\left(\frac{\pi_{\theta}(y|x)}{\pi_{\text{ref}}(y|x)}\right) + \beta \log Z(x)
\end{equation}
where $r_\theta(x, y)$ is an implicit reward function and $Z(x) = \sum_{y} \pi_{\text{ref}}(y|x) \exp\!\left(\tfrac{1}{\beta} r_\theta(x, y)\right)$ is the partition function ensuring normalization, which is intractable. Substituting this implicit reward formulation into the reward model training objective cancels out $Z(x)$, yielding the DPO loss
\[
L_{\text{DPO}}(\pi_\theta) = - \mathbb{E}_{(x, y_w, y_l) \sim D}\!\left[ \log \sigma\!\left(\beta \log \frac{\pi_{\theta}(y_w|x)}{\pi_{\text{ref}}(y_w|x)} - \beta \log \frac{\pi_{\theta}(y_l|x)}{\pi_{\text{ref}}(y_l|x)} \right) \right]
\]
where $(y_w, y_l)$ denote the preferred and dispreferred responses, respectively. Optimizing $L_{\text{DPO}}$ aligns the model directly with preference data without requiring an explicit reward model or a separate RL training loop. Consequently, DPO unifies the reward modeling and policy optimization stages of RLHF into a single, tractable procedure, greatly simplifying implementation and reducing memory cost.

Nonetheless, DPO introduces its own set of challenges. Since $Z(x)$ cannot be computed explicitly, DPO relies solely on pairwise preference data, rendering single-response supervision unusable during fine-tuning. This dependence on pairwise data--typically limited and expensive to collect--can lead to inefficient utilization of available feedback. Moreover, while RLHF's reward model can provide continuous-valued feedback for arbitrary prompts, DPO only learns from pairwise comparisons, limiting the granularity of optimization signals.

\subsection{UNA}
One limitation of DPO is that it can only utilize pairwise datasets, while other data formats cannot be leveraged. In addition, the pairwise preference data provide less information compared with the detailed information provided by a reward model. UNA extends DPO to pointwise datasets composed of (prompt, response, reward), i.e., $(x, y, r)$. In particular, UNA discovers that Equation~\ref{eq: appendix UNA reward} is also a special optimal implicit reward function for the RL objective in Equation~\ref{eq: appendix RL objective} based on DPO's implicit function in Equation~\ref{eq: appendix DPO equation}~\cite{wang2025unaunifyingalignmentsrlhfppo}:
\begin{equation}
\label{eq: appendix UNA reward}
\tilde{r}_{\theta}(x, y) = \beta \log\!\left(\frac{\pi_{\theta}(y|x)}{\pi_{\text{ref}}(y|x)}\right)
\end{equation}
Because the intractable $Z(x)$ has been removed, UNA can expand DPO to different types of signals rather than only pairwise signals. Eventually, the LLM policy is optimized through the MSE between implicit and explicit reward models as shown in Equation~\ref{eq: appendix UNA loss}.
\begin{equation}
\label{eq: appendix UNA loss}
L_{\text{UNA-reward}}(\pi_{\theta}) = \mathbb{E}_{(x,y) \sim D}\!\left[(r_{\phi}(x, y) - \tilde{r}_{\theta}(x, y))^2\right]
\end{equation}
UNA remains an offline method and lacks the on-policy exploration that RLHF, RLVR, and GIFT provide; this is the gap that GIFT closes by sampling on-policy and group-standardizing both the implicit and explicit rewards.

\section{Proof of Proposition~\ref{lem:cancellation}}\label{app:proof-cancellation}

The proof proceeds in three steps and shows that $z$-scoring the DPO implicit reward yields the same advantage as $z$-scoring the $\beta$-free implicit reward. The argument simply tracks how an additive constant and a multiplicative scalar transform under (i) the group mean, (ii) the group variance, and (iii) the resulting $z$-score--and the resulting advantage is invariant to both scalars.

\paragraph{Step 1 (Mean).} Recall the DPO implicit reward in Eq.~\ref{eq: DPO equation},
\(
r_\theta(x,y_i) = \beta \log\!\big(\frac{\pi_\theta(y_i|x)}{\pi_{\text{ref}}(y_i|x)}\big) + \beta\log Z(x) = \beta\,\hat r_\theta(x,y_i) + \beta\log Z(x)
\)
The term $\beta\log Z(x)$ depends only on $x$ and not on the response index $i$, so it is constant across the on-policy group. Averaging over the $N$ samples,
\begin{equation}\label{eq: DPO_mean}
\mu_\theta \;=\; \frac{1}{N}\sum_{i=1}^{N}r_\theta(x,y_i) \;=\; \beta\big(\frac{1}{N}\sum_{i=1}^{N}\hat r_\theta(x,y_i)\big) + \beta\log Z(x) \;=\; \beta\,\hat\mu_\theta + \beta\log Z(x)
\end{equation}
Subtracting the group mean from each $r_\theta(x,y_i)$ kills the additive constant $\beta\log Z(x)$ and leaves a clean rescaling of the centered $\beta$-free reward:
\begin{equation}\label{eq: GIFT_mean_subtraction}
r_\theta(x,y_i) - \mu_\theta \;=\; \beta\,\hat r_\theta(x,y_i) + \beta\log Z(x) - \beta\hat\mu_\theta - \beta\log Z(x) \;=\; \beta\big(\hat r_\theta(x,y_i)-\hat\mu_\theta\big)
\end{equation}

\paragraph{Step 2 (Variance).} Variance is invariant under additive shifts and scales by the square of any multiplicative factor: $\mathrm{Var}[c\,X+d] = c^2\mathrm{Var}[X]$ for any constants $c,d$. Applied to $r_\theta = \beta\hat r_\theta + \beta\log Z(x)$ with $c=\beta$ and $d=\beta\log Z(x)$,
\begin{equation}\label{eq: GIFT_variance}
\sigma_\theta^{2} \;=\; \mathrm{Var}\!\left[\beta\,\hat r_\theta + \beta\log Z(x)\right] \;=\; \beta^{2}\,\mathrm{Var}\!\left[\hat r_\theta\right] \;=\; \beta^{2}\,\hat\sigma_\theta^{2}
\end{equation}
so $\sigma_\theta = \beta\,\hat\sigma_\theta$.

\paragraph{Step 3 (Standardization).} Combining Eqs.~\ref{eq: GIFT_mean_subtraction} and~\ref{eq: GIFT_variance}, the group $z$-score of the DPO reward equals the group $z$-score of the $\beta$-free reward:
\begin{equation}\label{eq: GIFT Implicit Reward Normalization}
A_\theta(x,y_i)
\;=\; \frac{r_\theta(x,y_i)-\mu_\theta}{\sigma_\theta}
\;=\; \frac{\beta\big(\hat r_\theta(x,y_i)-\hat\mu_\theta\big)}{\beta\,\hat\sigma_\theta}
\;=\; \frac{\hat r_\theta(x,y_i)-\hat\mu_\theta}{\hat\sigma_\theta}
\;=\; \hat A_\theta(x,y_i)
\end{equation}
The factors of $\beta$ in the numerator and denominator cancel; the additive $\beta\log Z(x)$ is killed by the centering. Hence $A_\theta=\hat A_\theta$ pointwise on every group, and the GIFT loss is numerically identical whether one writes the DPO implicit reward with $\beta$ and $Z(x)$ or without them. \qed

\section{Proof of Theorem~\ref{thm:gift-optima}}\label{app:proof-thm-optima}

The proof rests on a single elementary fact about $z$-scores: two random variables have identical $z$-scores if and only if they are affine transforms of one another. The lemma is stated and proved first, then applied twice--once in each direction--to establish parts (a) and (b).

\begin{lemma}[$z$-score equality $\Leftrightarrow$ affine relation]\label{lem:zscore}
Let $X,Y$ be random variables on a common probability space with $\mathrm{Var}(X)>0$ and $\mathrm{Var}(Y)>0$. The following are equivalent:
\begin{equation*}
\textup{(i)}\;\;\frac{X-\mathbb{E}[X]}{\sqrt{\mathrm{Var}(X)}}\;=\;\frac{Y-\mathbb{E}[Y]}{\sqrt{\mathrm{Var}(Y)}}\quad\text{a.s.,}\qquad\textup{(ii)}\;\;\exists\,a>0,\,b\in\mathbb{R}:\;X=aY+b\quad\text{a.s.}
\end{equation*}
In (ii) the slope is uniquely $a=\sqrt{\frac{\mathrm{Var}(X)}{\mathrm{Var}(Y)}}$ and the intercept is $b=\mathbb{E}[X]-a\,\mathbb{E}[Y]$.
\end{lemma}
\begin{proof}
\emph{(ii) $\Rightarrow$ (i).} Assume $X=aY+b$ with $a>0$. By linearity of expectation, $\mathbb{E}[X]=a\,\mathbb{E}[Y]+b$, so $X-\mathbb{E}[X] = a(Y-\mathbb{E}[Y])$. Variance is invariant under additive shifts and scales by the square of any positive multiplicative factor, so $\mathrm{Var}(X)=a^2\mathrm{Var}(Y)$ and $\sqrt{\mathrm{Var}(X)} = a\sqrt{\mathrm{Var}(Y)}$. Substituting,
\[
\frac{X-\mathbb{E}[X]}{\sqrt{\mathrm{Var}(X)}} \;=\; \frac{a(Y-\mathbb{E}[Y])}{a\sqrt{\mathrm{Var}(Y)}} \;=\; \frac{Y-\mathbb{E}[Y]}{\sqrt{\mathrm{Var}(Y)}}
\]

\emph{(i) $\Rightarrow$ (ii).} Multiplying both sides of the $z$-score identity in (i) by $\sqrt{\mathrm{Var}(X)}$,
\[
X-\mathbb{E}[X] \;=\; \frac{\sqrt{\mathrm{Var}(X)}}{\sqrt{\mathrm{Var}(Y)}}\big(Y-\mathbb{E}[Y]\big)
\]
Setting $a=\sqrt{\frac{\mathrm{Var}(X)}{\mathrm{Var}(Y)}}>0$ and $b=\mathbb{E}[X]-a\,\mathbb{E}[Y]$, the right-hand side equals $aY+b-\mathbb{E}[X]$, so $X = aY+b$ a.s.
\end{proof}

\paragraph{Set-up.} Throughout, fix a prompt $x$ in the prompt dataset $D$ and view $r_\phi(x,\cdot)$ and $\hat r_\theta(x,\cdot)$ as random variables under $y\sim\pi(\cdot|x)$. Denote the means by $\mu_\phi(x)=\mathbb{E}_{y\sim\pi(\cdot|x)}[r_\phi(x,y)]$ and $\hat\mu_\theta(x)=\mathbb{E}_{y\sim\pi(\cdot|x)}[\hat r_\theta(x,y)]$. The two advantages are the corresponding $z$-scores: $A_\phi(x,y)=\frac{r_\phi(x,y)-\mu_\phi(x)}{\sigma_\phi(x)}$ and $\hat A_\theta(x,y)=\frac{\hat r_\theta(x,y)-\hat\mu_\theta(x)}{\hat\sigma_\theta(x)}$. The GIFT residual is the difference of these two $z$-scored rewards,
\[
\delta_\theta^\pi(x,y) = \hat A_\theta(x,y) - A_\phi(x,y) \;=\; \frac{\hat r_\theta(x,y)-\hat\mu_\theta(x)}{\hat\sigma_\theta(x)} - \frac{r_\phi(x,y)-\mu_\phi(x)}{\sigma_\phi(x)}
\]
and Assumption~\ref{ass:nondeg} guarantees that $\sigma_\phi(x)>0$ and $\hat\sigma_\theta(x)>0$, so the residual is well-defined.

\paragraph{Part (a): Sufficiency.} The aim is to show that the GRPO/RLHF optimum $\pi^{*}_\beta$ achieves $\delta_\theta^{\pi^{*}_\beta}(x,y)=0$ pointwise. Taking the logarithm of Eq.~\ref{eq: rlhf-path} (the closed-form RLHF optimum),
\[
\hat r_\theta(x,y)\;\big|_{\pi=\pi^{*}_\beta} \;=\; \log\!\frac{\pi^{*}_\beta(y|x)}{\pi_{\text{ref}}(y|x)} \;=\; \frac{1}{\beta}\,r_\phi(x,y) \;-\; \log Z_\beta(x)
\]
This is an affine function of $r_\phi(x,y)$ with positive slope $a = \frac{1}{\beta}$ and intercept $b=-\log Z_\beta(x)$. Lemma~\ref{lem:zscore} (direction (ii) $\Rightarrow$ (i)) then gives $z$-score equality of $\hat r_\theta(x,\cdot)$ and $r_\phi(x,\cdot)$ under $y\sim\pi^{*}_\beta$, so $\delta_\theta^{\pi^{*}_\beta}(x,y)=0$ for every $y\in\mathcal{Y}(x)$. Therefore $\mathcal{L}_{\text{GIFT}}(\pi^{*}_\beta)=\mathbb{E}_{x \sim D, y \sim \pi^{*}_\beta(\cdot | x) }\left[\left(\delta_\theta^{\pi^{*}_\beta}\right)^2\right]=0$.

\paragraph{Part (b): Necessity.} Suppose $\pi$ achieves $\delta_\theta^\pi(x,y)=0$ for all $y\in\mathcal{Y}(x)$, i.e.\ $\hat A_\theta(x,y)=A_\phi(x,y)$ pointwise. Applying Lemma~\ref{lem:zscore} (direction (i) $\Rightarrow$ (ii)) with $X:=\hat r_\theta(x,\cdot)$ and $Y:=r_\phi(x,\cdot)$, there exist $\alpha>0$ and $c\in\mathbb{R}$ with
\begin{equation}\label{eq: affine-step}
\hat r_\theta(x,y) \;=\; \alpha\,r_\phi(x,y) + c\qquad\text{for all }y\in\mathcal{Y}(x)
\qquad \alpha \;=\; \sqrt{\frac{\mathrm{Var}\,\hat r_\theta(x,\cdot)}{\mathrm{Var}\,r_\phi(x,\cdot)}} \;=\; \frac{\hat\sigma_\theta(x)}{\sigma_\phi(x)}
\end{equation}
Now use the definition $\hat r_\theta(x,y) = \log\frac{\pi_\theta(y|x)}{\pi_{\text{ref}}(y|x)}$ and exponentiate Eq.~\ref{eq: affine-step}:
\[
\pi_\theta(y|x) \;=\; e^{c} \pi_{\text{ref}}(y|x) e^{\alpha\,r_\phi(x,y)}
\]
The constant $e^c$ is determined by the requirement that $\pi(\cdot|x)$ be a probability distribution on $\mathcal{Y}(x)$:
\[
1 \;=\; \sum_{y\in\mathcal{Y}(x)}\pi(y|x) \;=\; e^c\sum_{y\in\mathcal{Y}(x)}\pi_{\text{ref}}(y|x)e^{\alpha\,r_\phi(x,y)} \;=\; e^c\cdot Z_{1/\alpha}(x)
\]
where the last equality uses the definition of $Z_\beta(x)$ in Eq.~\ref{eq: rlhf-path} with $\beta=\frac{1}{\alpha}$. Hence $e^c = \frac{1}{Z_{1/\alpha}(x)}$, and
\[
\pi(y|x) \;=\; \frac{\pi_{\text{ref}}(y|x)e^{\alpha\,r_\phi(x,y)}}{Z_{1/\alpha}(x)} \;=\; \pi^{*}_{1/\alpha}(y|x).
\]
Setting $\beta(x)=\frac{1}{\alpha} = \frac{\sigma_\phi(x)}{\hat\sigma_\theta(x)}$ yields Eq.~\ref{eq: beta-x}. Finiteness of $\beta(x)$ follows from Assumption~\ref{ass:nondeg} ensures $0<\beta(x)<\infty$.

\paragraph{Part (c): Loss minimization.} The pointwise residual $\delta_\theta^\pi(x,y)$ is real-valued, so $\left[\delta_\theta^\pi(x,y)\right]^2\ge 0$ and
\(
\mathcal{L}_{\text{GIFT}}(\pi) = \mathbb{E}_{x\sim D,\,y\sim\pi}[\left(\delta_\theta^\pi(x,y)\right)^2] \ge 0.
\)
Equality requires $\delta_\theta^\pi(x,y)=0$ on a set of full measure under $D\otimes\pi$. By Part~(a) every $\pi^{*}_{\beta(x)}$ satisfies $\delta_\theta^\pi(x, y)=0$ pointwise, and by Part~(b) only this family does. Hence $\mathcal{L}_{\text{GIFT}}(\pi_\theta)=0$ if and only if $\pi_\theta(\cdot|x)=\pi^{*}_{\beta(x)}(\cdot|x)$ for some $\beta(x)>0$ and $D$-almost every $x$. \qed

Parts (a) and (b) jointly say that GIFT's optimum is exactly the policies for which the implicit and explicit rewards have identical $z$-scores, and the only such policies are the GRPO/RLHF solution $\pi^{*}_{\beta(x)}(y|x)$. Part (b) further pins down the slope of the affine relation between $\hat r_\theta(x,y)$ and $r_\phi(x,y)$: it equals $\frac{\hat\sigma_\theta(x)}{\sigma_\phi(x)}$, and its reciprocal is the prompt-dependent inverse KL coefficient $\beta(x)=\frac{\sigma_\phi(x)}{\hat\sigma_\theta(x)}$ that GIFT implicitly selects.

\section{Concurrent and \texorpdfstring{$\beta$}-Free Preference Methods}\label{app:concurrent}

GIFT is not the first $\beta$-free preference objective, but it differs from prior $\beta$-free formulations on two axes: data regime (online vs.\ offline) and loss family (regression vs.\ classification or policy-gradient).

\paragraph{Offline $\beta$-free methods.} SimPO \cite{meng2024simposimplepreferenceoptimization} replaces the DPO log-ratio with a length-normalized reward margin $r_{\text{SimPO}}(x,y)=\frac{1}{|y|}\log\pi_\theta(y|x)$ and adds a fixed margin $\gamma$ to the BT loss; this removes $\beta$ but requires pairwise preference data. ORPO \cite{hong2024orpomonolithicpreferenceoptimization} adds an odds-ratio penalty
\(
-\log\sigma(\log\,\mathrm{odds}_\theta(y_w|x)-\log\,\mathrm{odds}_\theta(y_l|x))
\)
directly to the SFT loss, eliminating the reference policy and $\beta$ together but again requiring pairwise data. Both methods rely on a fixed offline preference dataset and inherit the corresponding distribution-shift issues.

\paragraph{On-policy preference methods.} 
RAFT \cite{dong2023raftrewardrankedfinetuning} generates multiple responses, filters out unsuccessful ones, and fine-tunes the policy using the remaining high-quality samples. Several subsequent methods further remove the KL divergence regularization term, thereby eliminating the need for the $\beta$ coefficient \cite{minimax2025minimaxm1scalingtesttimecompute, zhao2025geometricmeanpolicyoptimization, roux2025taperedoffpolicyreinforcestable}.
However, none of these uses an MSE between standardized implicit and explicit rewards. GIFT differs in two ways: (i) it uses regression for reward matching rather than reward maximization, and (ii) it group-$z$-scores both the implicit \emph{and} the explicit reward, which is what enables Theorem~\ref{thm:gift-optima}'s closed-form characterization.

\paragraph{Group-relative policy gradient methods.} DAPO and GSPO are recent extensions of GRPO that modify the clipping rule and the per-token vs.\ per-sequence weighting; despite these differences, all three methods remain policy-gradient approaches retaining $\beta$ as a hyperparameter, and GRPO and DAPO have shown strong performance in practice and they are utilized for comparison with GIFT.

\section{Worked Example: \texorpdfstring{$\beta(x)$} on a 3-Response Group}\label{app:worked-example}

To illustrate the variance-determined $\beta(x)$ concretely, consider a prompt $x$ with three responses $\{y_1,y_2,y_3\}$, explicit rewards $r_\phi=(0,1,2)$, and converged policy $\pi_\theta(\cdot|x)$ uniform with $\hat r_\theta=(\hat r_1,\hat r_2,\hat r_3)=(-1,0,1)$. Both quantities have group means $\mu_\phi=1$, $\hat\mu_\theta=0$ and group standard deviations $\sigma_\phi=\sqrt{\tfrac{2}{3}}$, $\hat\sigma_\theta=\sqrt{\tfrac{2}{3}}$. The group $z$-scores
\[
A_\phi = \tfrac{1}{\sqrt{\tfrac{2}{3}}}(-1,0,1) = \hat A_\theta
\]
are equal, so $\mathcal{L}_{\text{GIFT}}(\pi)=0$ and Theorem~\ref{thm:gift-optima} (b) yields $\beta(x)=\frac{\sigma_\phi(x)}{\hat\sigma_\theta(x)}=1$. The corresponding solution is $\pi^{*}_{1}(y|x)\propto\pi_{\text{ref}}(y|x)e^{r_\phi(x,y)}$. If the reward had been $r_\phi=(0,2,4)$ instead (the same ranking but doubled spread), $\sigma_\phi(x)$ would double while $\hat\sigma_\theta(x)$ stays fixed, so $\beta(x)$ would also double--confirming that $\beta(x)$ tracks reward dispersion endogenously. Conversely, if the policy converged to a sharper $\pi$ with $\hat r_\theta=(-2,0,2)$ on the same reward $(0,1,2)$, $\hat\sigma_\theta(x)$ would double, $\beta(x)$ would halve.

\section{Training Configuration}
\label{app:training-config}

All variants are trained on the DAPO math-17k corpus and evaluated on a three-benchmark suite
(AIME 2024, AIME 2025, AMC 2023) with
\texttt{rollout.n}=$32$, temperature $1.0$, and top-$p$\,$=0.7$, producing
mean@32, majority@32, and best@32 scores per checkpoint.
Table~\ref{tab:training-hparams} lists the per-variant hyperparameters
for both backbones.  Within each backbone, every variant shares the
same training/PPO batch sizes, rollout count, prompt/response length
caps, and entropy coefficient; only the KL coefficient (set to $1$
for GIFT and $0$ for the others) and the dynamic-sampling rollout
batch size differ across algorithms.

\begin{table}[ht]
  \centering
  \footnotesize
  \setlength{\tabcolsep}{4pt}
  \resizebox{\textwidth}{!}{%
  \begin{tabular}{llcccccccccc}
    \toprule
    Model & Variant & LR & \shortstack{train\\batch} & \shortstack{PPO mini\\batch} & \shortstack{rollout\\.n} & \shortstack{max prompt\\length} & \shortstack{max response\\length} & \shortstack{entropy\\coeff} & \shortstack{KL coef\\$\beta$} & \shortstack{rollout batch\\(dyn.\ sampling)} \\
    \midrule
    \multirow{4}{*}{\shortstack{Qwen3-30B\\-A3B-Base}}
      & GIFT  & $3\!\times\!10^{-6}$ & $1024$ & $256$ & $16$ & $2048$ & $8192$ & $0$ & $1$ & $3072$ \\
      & DAPO  & $3\!\times\!10^{-6}$ & $1024$ & $256$ & $16$ & $2048$ & $8192$ & $0$ & $0$ & $3072$ \\
      & GRPO  & $3\!\times\!10^{-6}$ & $1024$ & $256$ & $16$ & $2048$ & $8192$ & $0$ & $0$ & $1024$ \\
      & GSPO  & $3\!\times\!10^{-6}$ & $1024$ & $256$ & $16$ & $2048$ & $8192$ & $0$ & $0$ & $3072$ \\
    \midrule
    \multirow{4}{*}{\shortstack{Qwen3.5-35B\\-A3B-Base}}
      & GIFT  & $1\!\times\!10^{-6}$ & $1024$ & $256$ & $16$ & $2048$ & $8192$ & $0$ & $1$ & $3072$ \\
      & DAPO  & $1\!\times\!10^{-6}$ & $1024$ & $256$ & $16$ & $2048$ & $8192$ & $0$ & $0$ & $3072$ \\
      & GRPO  & $1\!\times\!10^{-6}$ & $1024$ & $256$ & $16$ & $2048$ & $8192$ & $0$ & $0$ & $1024$ \\
      & GSPO  & $1\!\times\!10^{-6}$ & $1024$ & $256$ & $16$ & $2048$ & $8192$ & $0$ & $0$ & $3072$ \\
    \bottomrule
  \end{tabular}}
  \caption{Training hyperparameters per (model, variant). The KL
  coefficient $\beta$ is set to $1$ for GIFT (consumed by the GIFT
  loss) and to $0$ for DAPO/GRPO/GSPO. The rightmost column reports
  the rollout batch size used by dynamic sampling
  (\texttt{gen\_batch\_size}\,$=3\times$\,\texttt{train\_batch\_size}
  for GIFT/DAPO/GSPO; $1\times$ otherwise).}
  \label{tab:training-hparams}
\end{table}



\end{document}